\documentclass[letterpaper, 10 pt, conference]{ieeeconf}  

\IEEEoverridecommandlockouts                              
\usepackage[utf8]{inputenc}
\usepackage[T1]{fontenc}

\let\labelindent\relax

\title{\LARGE \bf Predicting Parameters for Modeling Traffic Participants}

\usepackage{times}
\usepackage{multicol}
\usepackage{hyperref}
\usepackage{graphicx}
\usepackage{bm}
\usepackage[nolist]{acronym}
\usepackage{amsmath}
\usepackage{amssymb}
\usepackage[capitalise]{cleveref}
\usepackage{algorithm}
\usepackage[usenames,dvipsnames,table]{xcolor}
\usepackage{algorithmic}
\usepackage{booktabs}
\usepackage{array}
\usepackage{dblfloatfix}
\usepackage{tabularx}
\usepackage{enumitem}
\usepackage{subcaption}
\usepackage{diagbox}
\usepackage{etoolbox}\AtBeginEnvironment{algorithmic}{\small}
\usepackage{color,soul} 

\DeclareMathAlphabet{\pazocal}{OMS}{zplm}{m}{n}

\usepackage{lipsum}
\usepackage{color}
\usepackage[colorinlistoftodos]{todonotes}

\usepackage{lipsum}

\usepackage{amsfonts}
\usepackage{mathtools}

\usepackage{graphicx}

\newtheorem{theorem}{Theorem}[section]

\newtheorem{problem}[theorem]{Problem}

\usepackage[]{hyperref} 
\definecolor{darkred}{RGB}{150,0,0}
\definecolor{darkgreen}{RGB}{0,150,0}
\definecolor{darkblue}{RGB}{0,0,150}
\hypersetup{colorlinks=true, linkcolor=red, citecolor=blue, urlcolor=darkblue}

\usepackage{etoolbox}
\apptocmd{\thebibliography}{\scriptsize}{}{}

\author{
Ahmadreza Moradipari$^{1,2}$ 
\and Sangjae Bae$^1$ 
\and Mahnoosh Alizadeh$^2$ 
\and Ehsan Moradi Pari$^1$ 
\and David Isele$^1$ 
\thanks{
This work was funded by Honda Research Institute USA, Inc. $^1$Honda Research Institute USA, Inc., $^2$University of California, Santa Barbara.
Email: $\{$ahmadreza$\_$moradipari,alizadeh$\}$@ucsb.edu, 
$\{$ahmadreza$\_$moradipari,sbae,emoradipari,disele$\}$@honda-ri.com}%
}%

\begin{document}

\maketitle

\begin{abstract}
Accurately modeling the behavior of traffic participants is essential for safely and efficiently navigating an autonomous vehicle through heavy traffic. We propose a method, based on the intelligent driver model, that allows us to accurately model individual driver behaviors from only a small number of frames using easily observable features. On average, this method makes prediction errors that have less than 1 meter difference from an oracle with full-information when analyzed over a 10-second horizon of highway driving. We then validate the efficiency of our method through extensive analysis against a competitive data-driven method such as Reinforcement Learning that may be of independent interest.      
\end{abstract}


\IEEEpeerreviewmaketitle

\section{Introduction}
Generating safe and efficient behaviors for an autonomous driving (AD) agent requires an ability to model how other traffic participants will behave. Since the behavior of a traffic participant depends on the behavior of surrounding traffic agents, including our autonomous vehicle (ego vehicle), it must be conditioned on these surrounding agents. Moreover, because 
AD is a safety critical application, the behaviors must be robust to a large variety of circumstances.

Recently, data-driven methods have emerged with increasing volumes of dataset, appealing to both the automotive industry and academia with empirically proven performances under complex environments. In particular, variants of deep learning techniques have shown their effectiveness and computational efficiency for predicting AD environments \cite{choi2019drogon,li2020evolvegraph,wang2022transferable}. Although plausible and powerful, deep learning methods share technical and practical limitations that hinder the data-driven approaches being applied to real-world applications. The limitations are represented by interpretability (behind a choice of actions), generalizability (shifting domains), and the need of excessive data (for both general and corner cases), which are critical for safe-sensitive applications. Despite the growing set of the literature, deep learning methods require profound verification and validation until practically implemented. Therefore, simpler but easily implementable methods still remain as a popular choice of baseline in practice as well as research papers \cite{isele2019itsc,liu2020exploring,casas2021mp3}. 
%
%
In particular, the Intelligent Driver Model (IDM) \cite{treiber2000congested} has proven a valuable method for modelling traffic behavior and has been adapted to a wide variety of scenarios \cite{kesting2010enhanced,derbel2013modified,eggert2015foresighted,evestedt2016interaction,puphal2018optimization}. Estimating the behavioral parameters of IDM \cite{hoermann2017probabilistic} was proposed as a constrained learning problem for discovering the driver-specific behaviors of a traffic participant. This method provides accurate, individualized behaviors, while sufficiently constraining the model to prevent most erratic edge cases which could trigger undesirable reactions (i.e. phantom braking) from our ego agent. However there are some limitations to this approach: 1) acceleration is hard to measure accurately as it usually requires either differentiating velocity or double differentiating position which introduces noise to the measurement, 2) under normal driving conditions, times of relatively constant velocity are common, meaning one might not observe changes in velocity. This makes it difficult to accurately estimate an acceleration profile for a given driver. 
To address these issues we propose a method to \emph{predict} the IDM  parameters from other, more easily observable, traffic behaviors. Specifically, we investigate how accurately we can predict IDM parameters from lateral lane displacements, relative velocity, and the headway spacing of observed traffic participants.  
This allows us to quickly ascertain an individualized model for traffic participants over fairly long horizons.

The primary contribution of this work is presenting a prediction method for optimizing the use of a benchmark method for practical applicability and validating it through comparative analysis against a combination of heuristic and data-driven methods, with an accompanying analysis of the trade-offs of various aspects of the algorithm. This analysis includes a comparison of our heavily-constrained learning problem with a less-constrained learning-based approach to modeling behavior which might be of interest to the broader machine learning community. Specifically we observe how the two most commonly used metrics, average displacement error and final prediction error, when considered in isolation, can obscure the overall quality of an algorithm. While shortcomings of these metrics have been discussed before\footnote{https://towardsdatascience.com/why-ade-and-fde-might-not-be-the-best-metrics-to-score-motion-prediction-model-performance-and-what-1980366d37be}, our use-case provides a concrete example that many alternative metrics would also fail to recognize. While it has been shown that sometimes machine learning methods can be outperformed by much simpler methods \cite{scholler2020constant}, our results show that even when blackbox machine learning methods perform well according to quantitative analysis, they might have obvious failings not apparent from the metrics. This serves to highlight that thorough analysis should consider a diverse set of metrics and an assessment of broader practical issues (\emph{i.e} data, reliability, stability, interpretability) when deploying a system. While we believe machine learning has tremendous potential, our experiments serve to benchmark current capabilities and highlight the strengths and weaknesses of competing methods.

\section{Preliminaries}

Predicting the motion of other agents is essential when safely planning the trajectory for an autonomous vehicle. Understanding the uncertainty of a prediction has been one important line of research \cite{ma2020artificial}, but just knowing the uncertainty of predictions often lead to overly conservative behaviors which can cause unnecessary delays and make the ego agent's actions difficult for others to predict \cite{noh2018decision}. Deep learning techniques have been making huge strides in increasing the accuracy of prediction \cite{choi2019drogon,li2020evolvegraph,wang2022transferable} and will likely become the dominant method of prediction. However, for the time being, they suffer from distribution drift and often struggle to beat simple baselines like constant velocity \cite{scholler2020constant}, especially in cases like highway driving where constant velocity is strongly advantaged. For this reason, we turn to constrained learning problems, which while not as flexible as end-to-end learning, are easier to interpret, more robust to out-of-sample operation, and as a result are better able to fail gracefully and in easy-to-predict ways. 

In this paper we use IDM to constrain our learning problem. By estimating the parameters of IDM we are able to learn driver-specific models for traffic participants. However the IDM parameters, in practice, can be difficult to estimate accurately, so we present a method to predict IDM parameters from more easily measurable features. Before reviewing IDM.
 we discuss the notation we used in this work. 

Let $  {\mathcal{T} }= \{ \mathcal{X}_1,\dots,\mathcal{X}_T\}$ denote a sequence of the physical states for the modeled vehicle over a finite horizon $T$, and let $\mathcal{U} = \{u_1,\dots,u_T\}$, denote a finite sequence of 
control inputs. We denote $\mathcal{F}$ as the transition model such that for each time step $t$, we have $\mathcal{X}_{t+1} = \mathcal{F}(\mathcal{X}_{t}, u_{t+1})$. In this work, we  use the kinematic bicycle model from \cite{kong2015iv}. The detailed explanation of the transition model is presented in Section~\ref{sec:transition_model}.

\subsection{Intelligent Driver Model}

The intelligent driver model (IDM) \cite{treiber2000congested} was proposed to better understand the dynamics of traffic participants, and how changes in velocity contribute to traffic jams. IDM has grown into a popular choice for designing behaviors in simulation \cite{bouton2017belief,hubmann2018belief,hoel2019combining}, controlling autonomous agents \cite{derbel2013modified,eggert2015foresighted,bouton2020reinforcement}, and predicting traffic participants \cite{evestedt2016interaction,bhattacharyya2020online}. In IDM, the change in velocity $\dot{v}$ is described as
\begin{eqnarray} 
\dot{v} = a \left( 1-\left(\frac{v}{v_0}\right)^\phi - \frac{d^*(v,\Delta v)}{d}^2 \right) \enspace, \label{idm_velocity_model}
\end{eqnarray}
where 
\begin{eqnarray}\label{idm_params_equation}
d^*(v,\Delta v) = d_0 + d_1 \sqrt{\frac{v}{v_0}} + T v + \frac{v ~\Delta v}{2\sqrt{ab}} \enspace .
\end{eqnarray}

%
Here $v$ is the car's current velocity, $\Delta v$ is the difference in velocity with respect to the car in front, $d^*$ is the desired minimum gap distance,  
and $d$ is the actual gap distance to the vehicle in front. We assume the desired velocity $v_0$ is the road's speed limit, and following Treiber \cite{treiber2000congested} we keep the acceleration exponent $\phi$ fixed at $4$. This leaves five driver specific parameters: safe time headway $T$, maximum acceleration $a$, desired deceleration $b$, and 
jam distances $d_0$ and $d_1$. Note that Treiber identifies $d_1$ as being important for accurately modelling the differences in driving behavior. Together these parameters form the IDM behavior parameters $\theta_{\text{IDM}} = \{a,b,T,d_0, d_1\}$.

\subsection{Parameter Estimation}

It was shown that these IDM behavioral parameters can be estimated online using a particle filter \cite{hoermann2017probabilistic}; however, most of these parameters relate to acceleration which is typically measured as the first derivative of velocity or the second derivative of position, both of which are noisy. Moreover, under normal driving conditions, it is common that a car will maintain a roughly constant velocity, making it difficult to estimate the dynamic behavior.  To handle these issues, we propose a method to predict the IDM parameters for each traffic participant, from other more easily observable traffic behaviors. Here for each vehicle, we focus on the three observable features from the trajectory: (i) lateral lane displacement $\tau$; (ii) relative velocity $\nu$; (iii) headway spacing from the proceeding vehicle $\omega$. We note these features with the vector $\varphi = \{\tau, \nu, \omega \} \in \mathbb{R}^3$ and we refer to it as \textit{driving code}. Then, we propose a prediction method, that given the vector $\varphi$,  computes the IDM behavior parameters $\theta_{\text{IDM}}$ for each vehicle.
Next, we formally define the problem we seek to solve, then we present our prediction method. 

\section{Approach}
Our goal is to model the behavior of human drivers. To do this, we focus on the two dimensional continuous action space: acceleration and steering. Specifically, we first  compute the  acceleration and steering of the modeled vehicle at each time step, and then using the transition model in Section~\ref{sec:transition_model}, we can generate trajectories for traffic participants.  To model the acceleration of the human driven vehicle we use the IDM model in \eqref{idm_velocity_model}. This model, requires the IDM behavioral parameters, where we use a learning approach to predict the IDM parameters from their driving code. 
For steering we make the strong assumption that a car follows it's current lane, and to do this, we adopt the Pure Pursuit method from \cite{coulter1992implementation}. This assumption obviously makes errors whenever a car changes lanes, and provides room for improvement by less restricted models; however, we will show that with this restricted assumption, we gain a robustness that allows us to outperform more flexible models (see Section~\ref{sec:experiments} for details). The velocity and steering are then passed to transition model in \eqref{eq:kinematic_bicycle} to allow iteration of our models and rollout trajectories in simulation.

Two metrics we use in this work to measure the performance of our proposed algorithms and the baselines are average displacement error (ADE) and final displacement error (FDE). In particular, given ground truth trajectory, $\mathcal{T}^g$ expressed as a sequence of physical states $\mathcal{T}^g = \{\mathcal{X}_t\}_{t=1}^T$, we want to minimize the difference between the ground truth trajectory $\mathcal{T}^g$ and the trajectory generated by our model $\mathcal{T}^m$. Two popular metrics for measuring this difference are: 
\begin{align}
 &\text{ADE}(\mathcal{T}^g, \mathcal{T}^m) = \frac{1}{T}\sum_{t=0}^T \sqrt{ (x_t^m - x_t^{g})^2 + (y_t^m - y_t^{g})^2 }  \label{def:ADE}
 \\&
 \text{FDE}(\mathcal{T}^g, \mathcal{T}^m) = \sqrt{ (x_T^m - x_T^{g})^2 + (y_T^m - y_T^{g})^2 }, \label{def:FDE}
\end{align}
where $x_t$ and $y_t$ are Cartesian positions of the vehicles. These two metrics in isolation can obscure important qualities of the behaviors, so we also include a collision count in the performance evaluation.  Next, we formally state the problem we seek to solve in this work.
   
\subsection{Problem Definition}

We consider a human driver driving a car in a congested traffic scene. The goal is to accurately model the behavior of the driver so we can understand how the driver will respond to changes in the scene. Note that this is different from driving optimally as human drivers are in general not optimal, and an autonomous vehicle needs to interact with the participants on the road complete with whatever idiosyncrasies they might have. In the following we first express our prediction problem for learning the individualized IDM behavioral parameters in order to compute the acceleration at each time step, and then we present the method we adopt for computing the steering in Section~\ref{sec:steering}.

For each vehicle, we note its IDM parameters with the vector $\theta_{\text{IDM}} \in \mathbb{R}^5$ that models the velocity of the driver based on its current IDM state $\mathcal{S}_t^{\text{IDM}}$. 
The IDM state vector $\mathcal{S}_t^{\text{IDM}}$ consists of the car's current velocity, $v$, and relative speed $\Delta v$ with respect to the car in front, and gap distance $d_t$ i.e., $\mathcal{S}_t^{\text{IDM}} = \{ v_t, {\Delta v}_t, d_t \}$. Therefore, for each vehicle, given the IDM state $\mathcal{S}_t^{\text{IDM}}$ and  its behavioral parameter vector $\theta_{\text{IDM}}$, its acceleration can be expressed as  $\dot{v} = \texttt{IDM}_{\theta_{\text{IDM}}}(\mathcal{S}_t^{\text{IDM}})$. 
Next, we express the problem we are seeking to solve. 

\begin{problem}
\emph{Given a set of driving codes for each vehicle $\varphi$, can we learn a function  $f$  from expert data that maps $\varphi$ to accurate IDM behavioral parameters $\theta_{\text{IDM}}$?} 
Formally, we seek to solve the following feasibility problem such that for each vehicle given its ground truth $\mathcal{T}^g$ as well as feature vector $\varphi$:  
\begin{align} \label{problem:optimization}
    \min_{f} ~ \text{ADE}&(\mathcal{T}^g, \mathcal{T}^m)  \\
    \text{s.t.} ~ \theta_{\text{IDM}} &=  f(\varphi) \nonumber\\  
     ~~~~ u_t &= \texttt{IDM}_{\theta_{\text{IDM}}} (\mathcal{S}_t^{\text{IDM}}), \nonumber\\ 
     ~~~~ \mathcal{T}^m &= \{ \mathcal{F}_t(\mathcal{X}_t, u_{t+1}) \}_{t=0}^T \nonumber
\end{align}
\end{problem}
In particular, in our setting, given $\varphi \in \mathbb{R}^3$, we seek to find the  function $f: \varphi \rightarrow  \theta_{\text{IDM}}$ such that it minimizes the ADE in \eqref{def:ADE} of the modeled vehicle from its ground truth trajectory.

\subsection{Prediction Model}
For each vehicle $i$, we denote its IDM behavioral parameters and driving code with $\theta_{\text{IDM}}^i$ and $\varphi^i$, respectively. Our goal is that for the vehicle $i_*$, given its driving code $\varphi^{i_*}$ as well as  a set of IDM behavioral parameters and driving code  pairs from other traffic participants, $\Theta = \{ (\theta_{\text{IDM}}^1, \varphi^1),\dots,(\theta_{\text{IDM}}^N, \varphi^N )\}$, we predict an accurate IDM behavioral parameters $\theta_{\text{IDM}}^{i_*}$.   We employ the idea that the driving code $\varphi^j$, i.e., lateral lane displacement, relative velocity and headway spacing,  is a sufficient representative for the driving behavior of the vehicle $j$ in a highway. Because our set of driving codes $\varphi$ has a low dimension, we use nearest neighbor prediction, as this is known to work well for low dimensional problems \cite{pestov2013k}. We use the k-nearest neighbors (KNN) algorithm to find the $k$ closest driving codes from the set $\Theta$ to the $\varphi^{i_*}$. Then, we average their corresponding IDM behavioral parameters to compute $\theta_{\text{IDM}}^{i_*}$. 
Given $\theta_{\text{IDM}}^{i_*}$ we compute acceleration using \eqref{idm_velocity_model}.  

\subsection{Steering}\label{sec:steering}

Because IDM only provides the acceleration, we need a way to determine our steering for a complete control algorithm. Using the assumption that the vehicle will not change lanes, we can select steering that will keep the vehicle centered in the lane. This could be accomplished with a standard control algorithm like PID, but in practice we use a vector based version of pure pursuit \cite{coulter1992implementation} which preserves heading angle \cite{lundgren2003path} and is very easy to tune.

\subsection{Transition Model}\label{sec:transition_model}

In order to iterate a scene and generate     trajectories in simulation, we need to model the vehicle transitions. For this we use the kinematic bicycle model from \cite{kong2015iv}. Specifically, we define the physical state of the each vehicle at time $t$ as $ \mathcal{X}_t = (x_t, y_t, \psi_t, v_t )$ and we let $u_t = (\dot{v}_t, \delta_t)$ represents the control input, where $\dot{v}_t$ is the acceleration input and $\delta_t$ is the steering. Then we can write the transition model $\mathcal{F}_t(\mathcal{X}_t, u_{t+1})$ as:
\vspace{-0.75\baselineskip}
\small
\begin{align}\label{eq:kinematic_bicycle}
    \beta_k &= \text{tan}^{-1}\Bigg( \frac{l_r}{l_f+l_r} \text{tan}(\delta_k)\Bigg)  \\
    x_{k+1} &= x_k + v_k\text{cos}(\psi_k+\beta_k) \Delta t  \nonumber\\
    y_{k+1} &= y_k + v_k\text{sin}(\psi_k+\beta_k) \Delta t  \nonumber\\
    \psi_{k+1} &= \psi_k + \frac{v_k}{l_r}\text{sin}(\beta_k) \Delta t  \nonumber\\
    v_{k+1} &= v_k + a_k \Delta t, \nonumber
\end{align}
\normalsize
where $\psi$ is the heading angle, $l_r$ and $l_f$ are the distances from the center of the mass to the front and rear of the vehicle, respectively. 
This allows us to simulate traffic agents into the future using plausible transition dynamics. While we could simulate an entire traffic scene this way, for evaluation purposes we place one modeled agent in a pre-recorded scene, which we describe in the next section.

\section{Experiments}\label{sec:experiments}

\subsection{NGSIM data} 
We use the  Next Generation Simulation (NGSIM)\footnote{https://ops.fhwa.dot.gov/trafficanalysistools/ngsim.htm} dataset for US Highway 101. NGSIM provides 45 minutes of driving at 10 Hz. This dataset covers an area in Los Angeles approximately 640m in length. In our experiment, we focus on the five mainline lanes, and we remove the data for auxiliary lanes for the highway entrance and exit.  For training dataset, we consider the data from 7:50 a.m. to 8:05 a.m. which includes $1992$ cars with unique ID. For testing dataset, we consider the data from 8:05 a.m. to 8:20 a.m. which includes $1521$ cars with unique ID. In the test dataset, $8$ cars have frame ID irregularities, that cause the cars to disappear, and $29$ cars  have the wrong car in front at some frames, causing IDM to be blind to the true vehicle in front. These cars are removed from the dataset. Therefore, our test dataset has $1484$ cars with unique ID. 

\subsection{Baseline methods}
In this section, we compare the performance of our proposed algorithm with four other baseline methods on the test data. We report the average and standard error for the performance of the algorithms over a 10 seconds horizon (100 frames) in  Table~\ref{tab:performance_test_data}.  Because the traffic vehicles are recorded they cannot respond/react to realistic behaviors that differ from the original vehicle's behavior, and this can result in collisions. Since plausible but different behaviors are not inherently bad for a model, we differentiate these collisions from cases where the model vehicle is at fault. In our results, we report the collisions where the model vehicle is at fault.
Next, we describe the baseline methods.



\begin{table}[]
\caption{Test Results over 1484 cars}
\label{tab:performance_test_data}
\small
\begin{tabular}{cccc}
\hline
\toprule
\textbf{Method} & \textbf{ADE} & \textbf{FDE} & \textbf{Collisions} \\ \hline
Const. vel & 7.94 $\pm$0.19 & 14.36 $\pm$0.45 & 1467 \\
IRL + RL & 6.41 $\pm$0.32 & 17.50 $\pm$0.74 & 766 \\
IDM Ave. & 5.87 $\pm$0.17 & 8.94 $\pm$0.40 & 0 \\
IDM Pred. & 4.80 $\pm$0.15 & 7.40 $\pm$0.37 & 0 \\
IDM Est. (Oracle) & 4.38 $\pm$0.15 & 7.39 $\pm$0.36 & 0 \\
\bottomrule
\end{tabular}
\end{table}

\subsubsection{Constant Velocity}
While simple, a constant velocity method has been established as a good baseline approach for the straight driving in highways. In this method, the modeled car drives with a constant velocity with zero input action (i.e., with no acceleration and steering), and it does not receive any feedback from the environment. 


\subsubsection{IRL + RL}

\begin{figure}[t]
  \centering
  \includegraphics[width=0.45\linewidth]{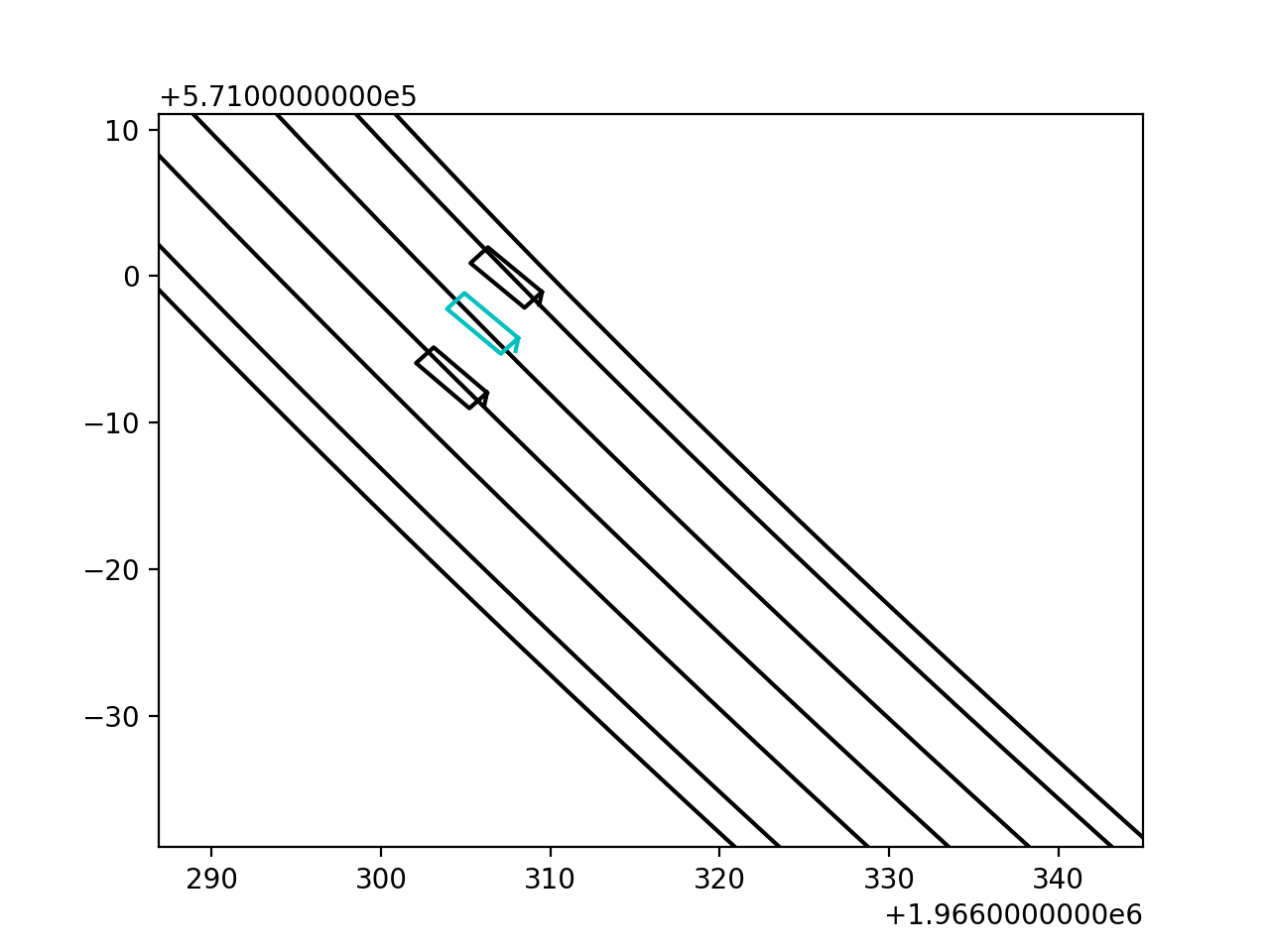} \\
  \includegraphics[width=0.5\linewidth]{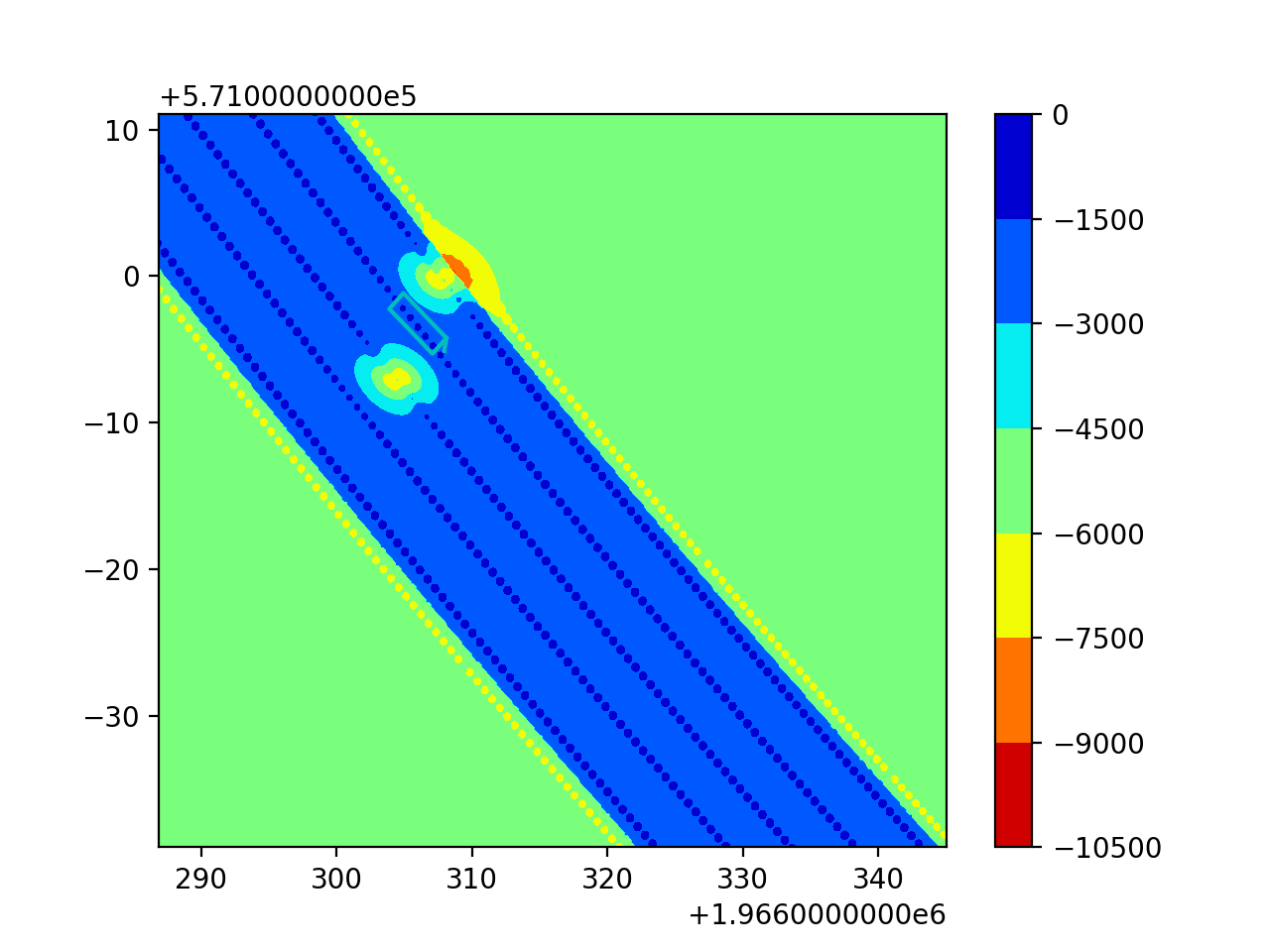}
  \caption{Features used in IRL from the modeled vehicle point of view. We show features corresponding to (i) holding center of the lanes, (ii) respecting the road boundaries, (iii) avoiding collisions with the other cars. }
  \label{reward_heat_map}
\end{figure}

To have a less constrained learning agent, we learn behaviors of a traffic agent using 
Inverse Reinforcement Learning (IRL). In our implementations, we learn the reward function from human data. We collect trajectories of human (expert) drivers from the NGSim dataset, and use IRL to recover a reward function that explains the trajectories.  To handle continuous state and action space, we employ continuous inverse optimal control \cite{levine2012continuous}. We model the reward function as a linear combination of predefined features and learn the reward weights corresponding to each feature. Then, we apply the principle of maximum entropy \cite{ziebart2008maximum} to optimize the reward weights in order to make the human demonstrations more likely. 
We display the heat map of the features we used in Figure \ref{reward_heat_map}.  The features we include in the reward function are as follows: 

\begin{itemize}
    \item distance to the middle of the lane
    \item distance to the boundaries of the road
    \item higher speed for moving forward 
    \item heading: in order to align the heading of the vehicle along with the road direction
    \item collision avoidance: we define a non-spherical Gaussian distribution on each car using its positions and heading  to compute the probability of collision to the other cars. The formulation of the non-spherical Gaussian is included in the Appendix~\ref{appendix:Gaussian}.
\end{itemize}

Then, using the learned reward function, we train a Proximal Policy
Optimization (PPO) algorithm \cite{schulman2017proximal} as our reinforcement learning agent. We trained the PPO agent using the training data over 1992 cars each with 10 seconds (100 frames) horizon. During the training process, we penalized the learning agent for colliding and driving off the road boundaries in order to reduce the state space exploration of the learning agent away from bad states. After training, we evaluate the performance of the trained model on the test data over 1484 cars for 10 seconds horizon. We report the results in \ref{tab:performance_test_data}.

\begin{figure*}[t]
  \centering
  \includegraphics[width=0.32\linewidth]{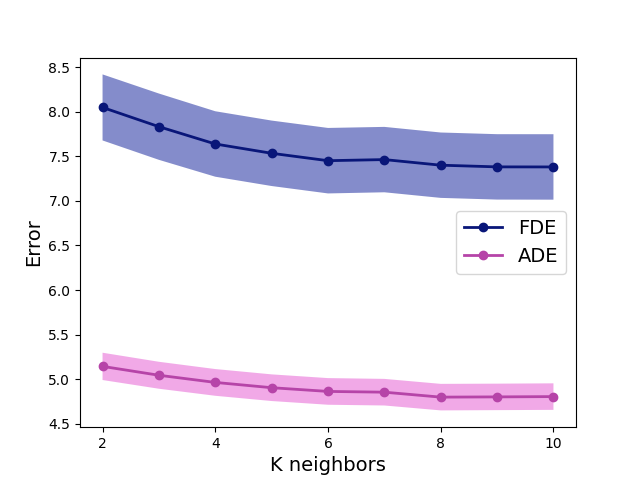}
  \includegraphics[width=0.32\linewidth]{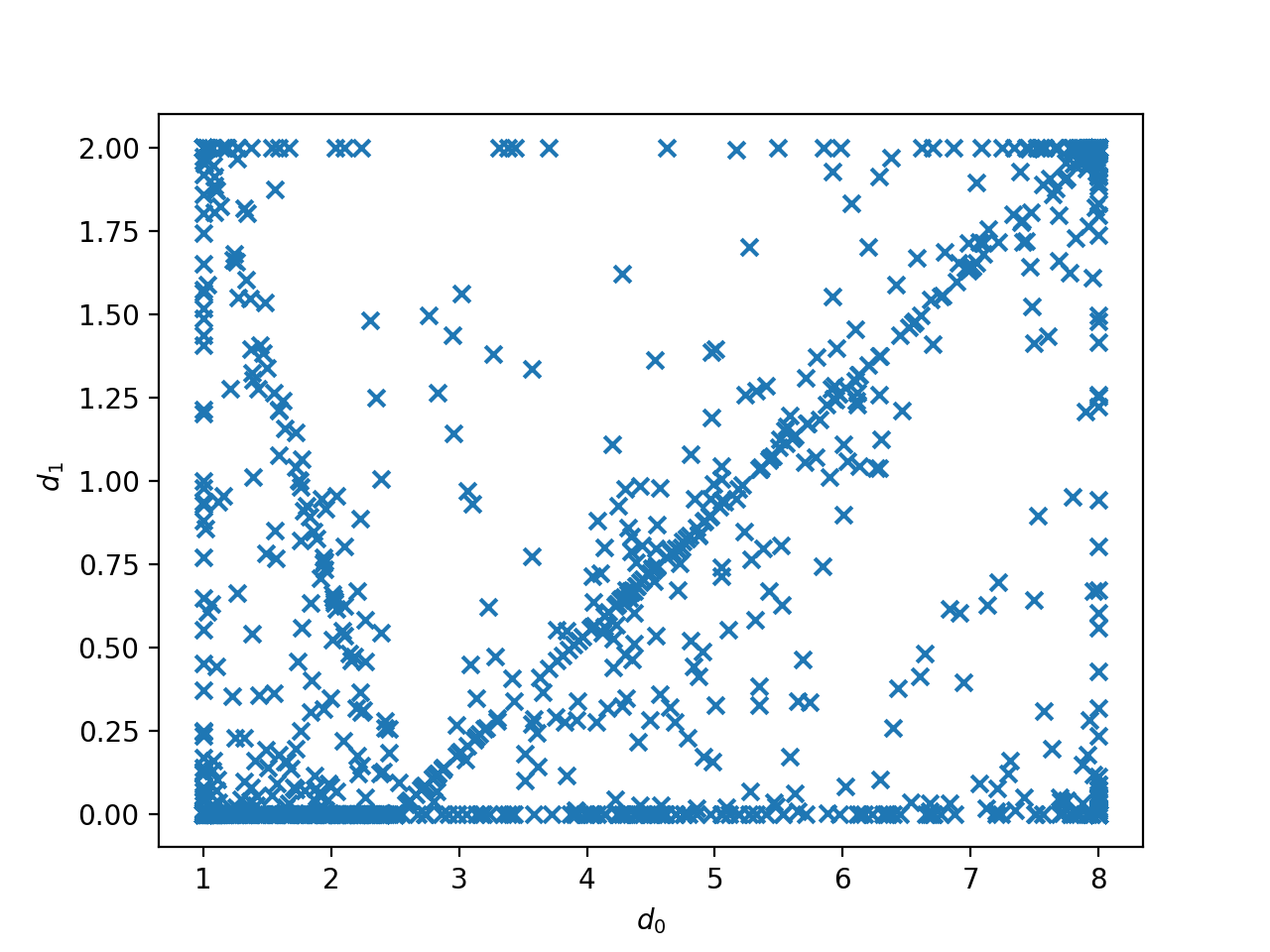}
  \includegraphics[width=0.25\linewidth]{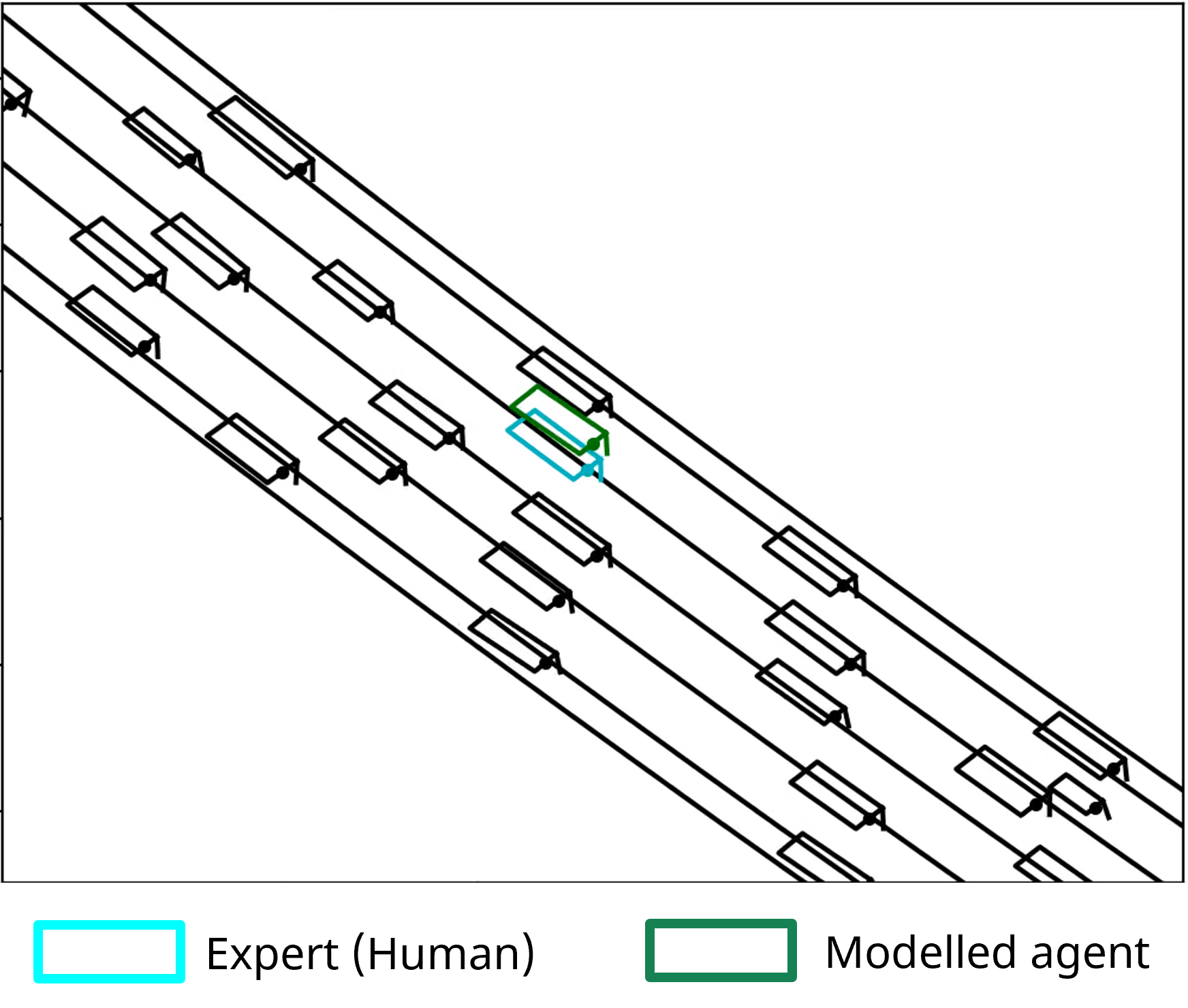}
  \caption{Left: Effect of K on model accuracy. Middle: A visualization of model parameters learned by the offline optimization. Note the parameter space is not uni-modal. Right: A visualization of the behavior or IRL. The modeled car drifts frighteningly close to other traffic participants. While this particular case did not lead to a collision, it is unnatural driving behavior that poorly models the true traffic participant.}
  \label{fig:differen_k_idm_dist}
\end{figure*}

\subsubsection{IDM Estimation (oracle)}

Here, we assume that the oracle has access to the full trajectory in the dataset (i.e., full-information method), and it estimates the IDM parameters using "L-BFGS-B" method in order to minimize the euclidean distance of the modeled car from the expert (ADE). In particular, we assume that for each expert in the training (or testing) dataset, the oracle has access to the entire 10 seconds (i.e., 100 frames) of the expert's trajectory, and it computes IDM parameters $\theta_{IDM}^{\text{train}}$ (or $\theta_{IDM}^{\text{test}}$ ) to learn driver-specific models for traffic participants.  Then, we evaluate the performance of the optimized IDM parameters from the full-information method for the 10 seconds horizon and we report the results in  Tables~\ref{tab:performance_test_data}. 

\subsubsection{IDM Average}
To confirm that our prediction method is predicting something meaningful, we compare our approach against the average of the IDM parameters from the training data. We report its result in the Table~\ref{tab:performance_test_data}. 
Since our prediction approach outperforms the average model, it indicates that the parameter space has meaningful variation. To verify this, we visualize two of the parameters $d_0 $ and $d_1$ of \eqref{idm_params_equation} in figure~\ref{fig:differen_k_idm_dist} (middle). Note that \cite{treiber2000congested} identified $d_1$ as being important for modeling differences in driver behaviors.

\subsubsection{IDM Prediction}

IDM prediction is our proposed approach. Given the trajectories of the experts, we create a vector of driving code for each vehicle  representing its driving style, $\varphi$. For each expert in the training dataset, the driving code is a vector of three dimensions that consists of the average of  (i) velocity, (ii) offset from the main lane (iii), headway time that represents the time to travel from the expert vehicle to the proceeding vehicle,  over the whole trajectory. We compute the driving code $\varphi$ for experts in the test dataset, given the first 10 frames (1 second) of their trajectories. Then, for each expert in the test dataset, we use KNN to find the $k$ closest driving codes from the set of $\Theta^{\text{train}}$ from the training experts. Then, we averaged their corresponding IDM parameters (computed from the full-information algorithm on the training dataset) in order to compute our prediction of the IDM parameters for the experts in the test dataset. We note that this approach requires less information and is less computationally expensive than the full-information approach. 
On average, predicting the IDM behavioral parameters for each car in the test dataset, takes $0.001$ seconds that shows its real-time applicability. We report the performance of the IDM prediction in Tables~\ref{tab:performance_test_data}.
Moreover, in Table~\ref{tab:driving_code-performance}, we show the effect of each combination of the driving code features on the performance of the IDM prediction method on the test data. 
In our prediction approach, we assume that we have access to the first 1 seconds (10 frames) of the trajectories in the test dataset. One might ask how robust is this prediction approach to the available information of the test dataset? To answer this, we evaluate the performance of our prediction approach using different number of available frames from the test data. We report the results in Table~\ref{tab:belief}.
Moreover, in Figure~\ref{fig:differen_k_idm_dist} (left), we show the effect of the number of neighbors in KNN on the performance of our prediction approach. Then, we select $K=8$ for the rest of our experiments.


\begin{table}
\caption{Effect of Driving Code on IDM Pred.}
\label{tab:driving_code-performance}
\small\centering
\begin{tabular}{cccc}
\hline
\toprule
\textbf{Method} & \textbf{ADE} & \textbf{FDE} & \textbf{Collisions} \\
\hline
$\varphi = \{\nu\}$   & 5.57 $\pm$0.17   & 8.50 $\pm$0.39  & 0 \\
$\varphi = \{\omega\}$     & 4.90 $\pm$0.15   & 7.60 $\pm$0.37   & 0 \\
$\varphi = \{\tau\}$     & 5.35 $\pm$0.17   & 8.11 $\pm$0.38   & 0 \\ 
$\varphi = \{\nu,\omega\}$     & 4.95 $\pm$0.15   & 7.60 $\pm$0.37   & 0 \\ 
$\varphi = \{\nu, \tau\}$ & 5.53 $\pm$0.16  & 8.34 $\pm$0.39   & 0 \\ 
$\varphi = \{\omega, \tau\}$ & 4.76 $\pm$0.15  & 7.35 $\pm$0.37   & 0 \\ 
$\varphi = \{\nu, \omega, \tau\}$ & 4.80 $\pm$0.15  & 7.40 $\pm$0.37   & 0 \\ 
 \bottomrule
\end{tabular}
\end{table}

\begin{table}
\caption{Effect of Frame Number}
\label{tab:belief}
\small\centering
\begin{tabular}{cccc}
\hline
\toprule
\textbf{Frame Number} & \textbf{ADE} & \textbf{FDE} & \textbf{Collisions} \\ \hline
2               & 5.32 $\pm$0.15  &  8.31 $\pm$0.37  & 0 \\
4               & 5.22 $\pm$0.15   & 8.23 $\pm$0.37   & 0 \\ 
6               & 5.01 $\pm$0.15   & 8.05 $\pm$0.37   & 0 \\ 
10              &  4.80 $\pm$0.15 &  7.40 $\pm$0.37  & 0 \\
20              & 4.80 $\pm$0.15    & 7.40 $\pm$0.37   & 0 \\
 \bottomrule
\end{tabular}
\end{table}

\subsection{Discussion}

In Tables~\ref{tab:performance_test_data} we observe that the our IDM parameter prediction method is within half of a meter of the full optimization on the test data. Note that this is at high speed after a relatively long horizon of 10 seconds. Currently many deep learning prediction dataset predict less than 5 second horizons \cite{li2020evolvegraph}.

We observe approximately a meter of prediction improvement over the averaged IDM parameters. This suggests the parameter space cannot be modelled well with a fixed set of parameters. Figure~\ref{fig:differen_k_idm_dist} (middle) plots the models learned from the full-information process. We observe multiple linear trends, we observed similar trends in the other feature dimensions (not shown for space considerations). The clustering around the edges is caused by the optimization bounds. 
In Table~\ref{tab:driving_code-performance}, we observe that  the IDM Prediction method considering only a relative velocity feature as a driving code, has a performance improvement over the  IDM average method. Also, we observe that headway spacing and lateral lane displacement features are sufficient for the IDM prediction to achieve a less than half a meter error in the longitude with respect to the oracle in a 10 seconds horizon.   

Additionally, we observe that RL,  outperforms constant velocity on the test dataset. This is because RL benefits from having a learned reward model - as has been discussed in prior literature \cite{osa2018algorithmic}, explicitly modelling a reward function enables an agent to generalize better than methods like behavior cloning \cite{pomerleau1988alvinn}. Because RL is less constrained it is able to accomplish many maneuvers not available to IDM, lane changes being one notable example. However this comes at the cost of RL taking much more risky behaviors and often colliding with traffic participants. We also observe that RL has a much greater variance than other methods. The extent of this is very pronounced as there are times when RL is right on top of the traffic agent for the whole run, and times when it speeds off through traffic like an aggressive motorcyclist. 

With the straightforward implementability, leveraging the proposed method could benefit the AD community in various aspects. Examples include (i) enhancing a simulation environment with more realistic settings for testing AD, comprehending the diversity of drivers' characteristics, (ii) enhancing baseline prediction modules (e.g., replacing naive constant velocity model) while not requiring an ample dataset or complex vehicle dynamics, and (iii) integrating it into a planning algorithm, accommodating the real-time applicability of IDM.

One limitation is associated with the validation environment which neglects the inter-agent interactions. Specifically, the trained models (both IDM and RL) are tested by rolling out the behaviors in the (NGSim) data-set offline, without empowering traffic participants to react to the behavior changes of the trained vehicle. This is done for standardization, \emph{i.e.} IDM might not collide with other IDM agents, but that doesn't mean IDM would be safe with IRL agents. By using recorded behavior, modelled agents are prevented from appearing safe because of other traffic agent's adjustments. It does, however, mean any benefits of interactivity are ignored, possibly to the agents detriment. Another limitation is inherited from the fact that IDM does not model a lane-changing behavior. Thus, IDM is often used together with a standalone lane changing model (e.g., MOBIL \cite{kesting2007general}). This modification is applicable to our proposed method.

\section{Conclusion}
In this work, we study the problem of learning/predicting driver-specific models to accurately model the behavior of traffic participants.  We constrained this learning problem based on intelligent drive model (IDM),  and we propose a prediction method that is able to efficiently learn the individualized IDM behavioral parameters from human data.  We then compare the performance of our proposed method with other less-constrained data-driven methods such as reinforcement learning (RL) whose reward function has been learned from human data using inverse reinforcement learning (IRL). 

\bibliographystyle{IEEEtran}

\bibliography{irlmpc}

\section*{Appendix: Ellipsoidal Gaussian Risk Measure}\label{appendix:Gaussian}

\begin{figure}[ht]
  \centering
  \includegraphics[width=0.8\linewidth]{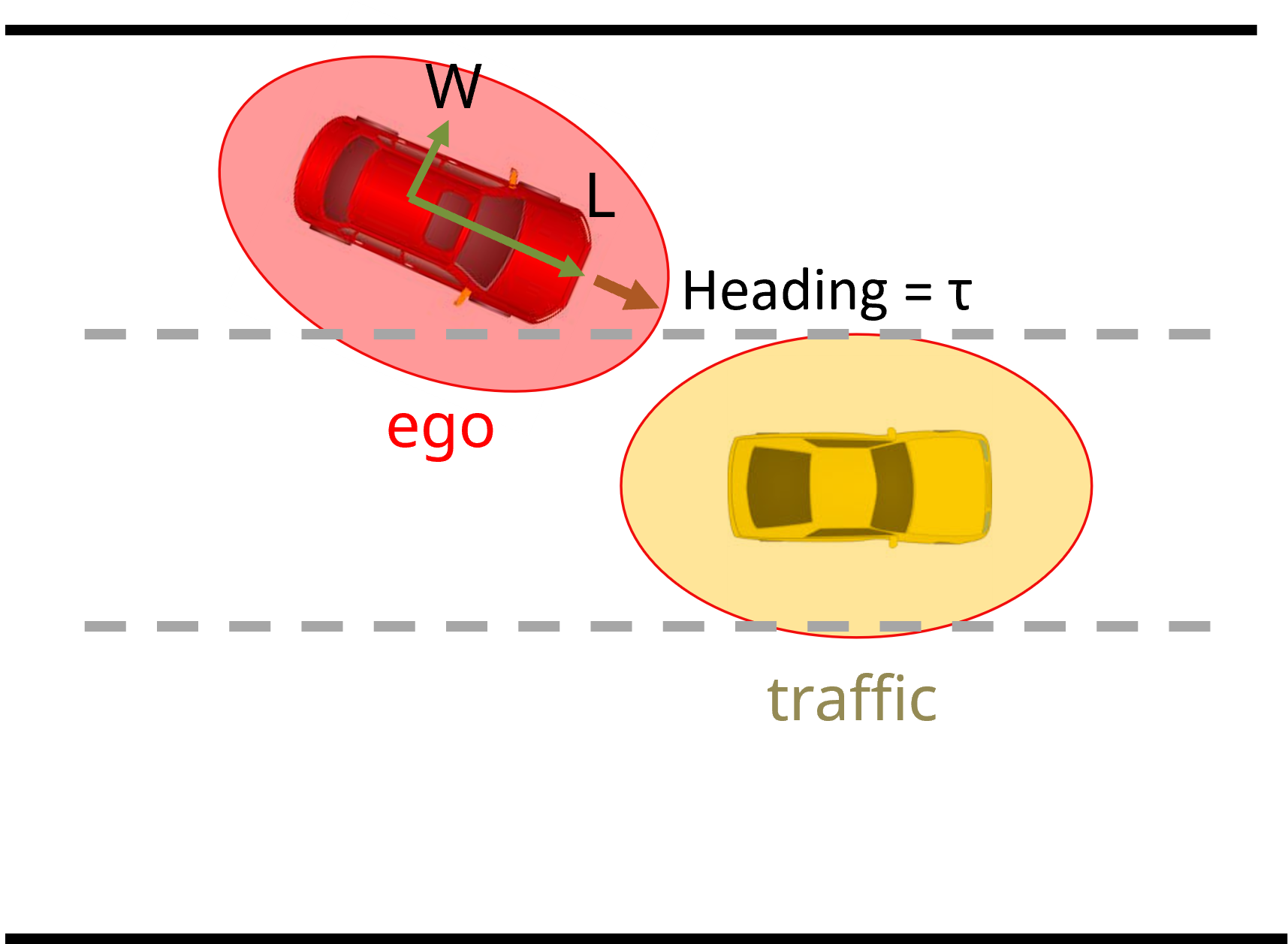}
  \caption{Ellipsoidal Gaussian Risk Measure}
  \label{fig:differen_k}
\end{figure}
We define an ellipsoidal Gaussian for the ego vehicle $\mathcal{N}(\mu, \Sigma)$, where $\mu$ is the $\{x,y\}$ position of the vehicle. 

\begin{gather}
 \Sigma
 =
  \begin{bmatrix}
   cos(\tau) & -sin(\tau) \\
   sin(\tau) & cos(\tau) 
   \end{bmatrix}
   \begin{bmatrix}
   L & 0 \\
   0 & W 
   \end{bmatrix}
   \begin{bmatrix}
   cos(\tau) & sin(\tau) \\
   -sin(\tau) & cos(\tau) 
   \end{bmatrix}
\end{gather}

Similarly, we define the traffic vehicle as $\mathcal{N}(\theta, \Gamma)$. The risk is then defined:
\begin{equation} \label{eq:g}
   \text{ risk } = \int \int \mathcal{N}(\mu, \Sigma) \enspace \mathcal{N}(\theta,\Gamma) \enspace.
\end{equation}

We let $\eta$ be a two dimension vector of $\{x,y\}$ grid points. We can then rewrite \ref{eq:g} as:
\begin{eqnarray}
   \text{ risk } = \frac{1}{(2\pi)^2 \sqrt{|\Sigma| |\Gamma| } }
  \int \int e^{\frac{-1}{2} \kappa}\\
  \kappa = (\eta-\mu)^\top \Sigma^{-1} (\eta-\mu) + (\eta-\theta)^\top \Gamma^{-1} (\eta-\theta)  
\end{eqnarray}
Multiplying and combining terms we get:
\begin{align*}
  \kappa &= \eta^\top(\Sigma^{-1}+\Gamma^{-1})\eta 
            - \eta^\top \Sigma^{-1} \mu 
            - \mu^\top \Sigma^{-1} \eta
            + \mu^\top \Sigma^{-1} \mu
            - \eta^\top \Gamma^{-1} \theta 
            - \theta^\top \Gamma^{-1} \eta
            + \theta^\top \Gamma^{-1} \theta \\
         &= \eta^\top(\Sigma^{-1}+\Gamma^{-1})\eta 
            - 2 \eta^\top (\Sigma^{-1}\mu+\Gamma^{-1}\theta)
            + \mu^\top \Sigma^{-1} \mu
            + \theta^\top \Gamma^{-1} \theta \enspace.\\
\end{align*}

We then define: 
\begin{align}
    \Omega &=\left( \Sigma^{-1} + \Gamma^{-1}\right)^{-1} \\
    v &=\Omega\left(\Sigma^{-1}\mu+\Gamma^{-1}\theta\right) \enspace. 
\end{align}

This results in a new form:
\begin{align*}
  \kappa &= \eta^\top \Omega^{-1} \eta
            - 2\eta^\top \Omega^{-1}) v
            + \mu^\top \Sigma^{-1} \mu
            + \theta^\top \Gamma^{-1} \theta \\
         &= (\eta-v)^\top \Omega^{-1} (\eta-v)
            - v^\top \Omega^{-1}) v
            + \mu^\top \Sigma^{-1} \mu
            + \theta^\top \Gamma^{-1} \theta \\
\end{align*}

Next, since we know the integral of a Gaussian PDF:
\begin{eqnarray}
\frac{1}{2\pi \sqrt{|\Omega| } }
  \int \int e^{\frac{-1}{2} (\eta-v)^\top \Omega^{-1} (\eta-v)}
  = 1 \enspace,
\end{eqnarray}
the final analytic form for the risk of two Gaussian distributions, $\mathcal{N}(\mu, \Sigma) \text{ and } \mathcal{N}(\theta,\Gamma)$, can be written as:

\begin{equation}
    \text{risk} = \frac{\sqrt{|\Omega|}}{2\pi\sqrt{|\Sigma| |\Gamma|}} e^{\frac{1}{2}\left(v^\top\Omega^{-1}v-\mu^\top\Sigma^{-1}\mu-\theta^\top\Gamma^{-1}\theta\right)} \enspace .
\end{equation}

\end{document}